\documentclass{article}

\usepackage{arxiv}

\usepackage[utf8]{inputenc} %
\usepackage[T1]{fontenc}    %
\usepackage{hyperref}       %
\usepackage{url}            %
\usepackage{booktabs}       %
\usepackage{amsfonts}       %
\usepackage{nicefrac}       %
\usepackage{microtype}      %
\usepackage{xcolor}         %
\usepackage{subfigure}
\usepackage{graphicx}
\usepackage{multicol}
\usepackage{multirow}
\usepackage{etoolbox}
\usepackage{xurl}

\title{Array Camera Image Fusion using Physics-Aware Transformers}

\author{
    {Qian Huang}\thanks{Qian Huang and Minghao Hu are students at Duke University. This work was finished when they were doing internship at the University of Arizona.} \\
	Wyant College of Optical Sciences\\
	University of Arizona\\
	Tucson, AZ 85721 \\
	\texttt{qh38@arizona.edu} \\
	\And
    {Minghao Hu} \\
	Wyant College of Optical Sciences\\
	University of Arizona\\
	Tucson, AZ 85721 \\
	\texttt{mh432@arizona.edu} \\
	\And
    \href{https://orcid.org/0000-0001-5655-2478}{\includegraphics[scale=0.06]{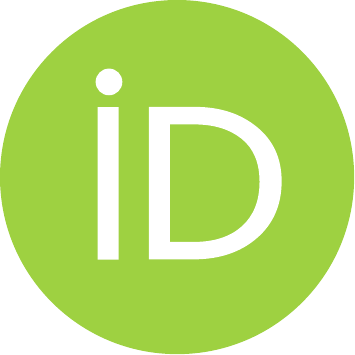}\hspace{1mm} David J. Brady} \\
	Wyant College of Optical Sciences\\
	University of Arizona\\
	Tucson, AZ 85721 \\
	\texttt{djbrady@arizona.edu}
}

\begin{document}
\maketitle

\begin{abstract}
We demonstrate a physics-aware transformer for feature-based data fusion from cameras with diverse resolution, color spaces, focal planes, focal lengths, and exposure. We also demonstrate a scalable solution for synthetic training data generation for the transformer using open-source computer graphics software. We demonstrate image synthesis on arrays with diverse spectral responses, instantaneous field of view and frame rate.
\end{abstract}

\section{Introduction}
In contrast with systems that use physical optics to form images, computational imaging uses physical processing to code measurements but relies on electronic processing for image formation~\cite{mait2018computational}. In optical systems, computational imaging enables "light field cameras~\cite{ihrke2016principles}" that capture high dimensional spatio-spectral-temporal data cubes.
The primary challenge of light field camera design is that, while the light field is 3, 4 or 5 dimensional, measurements still rely on 2D photodetector arrays. High dimensional light field capture on 2D detectors can be achieved using three sampling approaches: interleaved coding, temporal coding and multiaperture coding. Interleaved coding, as is famously done with color filter arrays~\cite{lukac2005color}, consists of enabling adjacent pixels on the 2D plane to access different parts of the light field. Mathematically similar sampling strategies for depth of field are implemented in integral imaging~\cite{xiao2013advances} and plenoptic cameras~\cite{ng2005light}. This interleaved approach generalizes to arbitrary high dimensional data cubes in the context of snapshot compressive imaging~\cite{yuan2021snapshot}. Temporal coding consists of scanning the spectral~\cite{hu2019multispectral} or focal~\cite{wang2021deep} response of the camera during recording.  

Array cameras~\cite{brady2018parallel} offer many potential advantages over interleaved and temporal coding. The advantage over temporal scanning is obvious, a camera array can capture snapshot light fields and does not therefore sacrifice temporal resolution. Additionally, development of cameras with dynamic spectral, spatial and focal sampling is more challenging than development of array components that sample slices of the data cube. The advantage of multiaperture cameras relative to interleaved sampling is more subtle, although implementation of interleaved sampling is also physically challenging. On a deeper level, however, interleaved sampling makes the physically implausible assumption that temporal sampling rates and exposure should be the same for different regions of the light field. In practice, photon flux in the blue is often very different from in the red and setting these channels to a common exposure level is unfortunate. The design of lenses and sensors optimized for specific spectral and focal ranges leads to higher quality data. 

With these advantages in mind, many studies have previously considered array cameras for computational imaging~\cite{tanida2016multi, plemmons2007periodic, Shankar:06}. More recently, artificial neural networks have found extensive application in array camera control and image processing~\cite{brady2020deep, yuan2021modular}. Of course, biological imaging systems rely heavily on array imaging solutions. While conventional array cameras originally relied on image-based registration~\cite{juan2010surf} for "stitching", biological systems integrate multiaperture data deep in the visual cortex.
In analogy with the biological system approach, here we demonstrate that array camera image fusion from deep layer features, rather than pixel maps, is effective in data fusion from diverse camera arrays. Our approach is based on transformers~\cite{vaswani2017attention, dosovitskiy2020image} networks, which excel at establishing long-range connections and integrating related features.

Since transformer networks are more densely connected than convolutional networks, high computational costs have inhibited their use in common computer vision tasks. Non-local neural connections drastically increase the receptive field for each voxel. As shown here, however, when the transformer is integrated with the physics of the system the connections outside of the physical receptive fields can be trimmed to the extent that the complexity of transformers is comparable to convolutional networks.

As with many physical image capture and processing tasks, the forward model for array camera imaging is easily simulated but the inverse problem is difficult to describe analytically. This class of problems can be addressed by training neural processors on synthetic data. Recent achievements in render engines like GPU-aided ray tracing allow us to realistically, accurately and efficiently model the world and render the modeled world to sensors of ideal virtual cameras. The rendered frames can be regarded as the ground truth. Synthetic sensor data that is degraded can be generated from the ground truth via the forward model of the camera array.
Surprisingly, it is unnecessary to build photorealistic real-world scenes for some computer vision problems, including image fusion; scenes with abstract objects native to computer graphics software with diverse colors and textures can span the problem domain of image fusion. Along with the programming interface provided by Blender~\cite{blender}, this data synthesis pipeline can be easily deployed and automatically scaled up to fit diverse data demands.

Here we use this approach to build a physics-aware transformer (PAT) network that can fuse data from the array cameras of the diverse resolution, color spaces, focal planes, focal lengths, and exposure. %
The purpose of this system is to combine data from array cameras to return a computed image superior to the image available to any single camera. Array cameras are designed in these systems to exploit differences in the spatial and temporal resolution needed to capture color, texture and motion. 

We demonstrate four example systems. The first system combines data from wide field color cameras with narrow field monochrome cameras, the second combines color information from visible cameras with the textural information from near infrared cameras, the third combines short exposure monochrome imagery with long exposure narrow spectral band data and the fourth combines high frame rate monochrome data with low frame rate color images. As a group, these designs combine with PAT processing to show that array cameras optimized for effective data capture can create virtual cameras with radically improved {dynamic range}, color fidelity and spatial/temporal resolution. 

\section{Related Work}

There are three main branches of combining physical models with neural algorithms. First, plugging a learned model as a prior into a physical model, which is also known as "plug and play". RED~\cite{romano2017little} is an example that applied a denoiser as its prior. The second way initiated by Deep image priors~\cite{ulyanov2018deep} is using a network architecture as a prior, thus removing the requirement of pretraining. The methods above, however, are optimized for a scene in a loop, 
restricting real-time applications. The third method is integrating the physics of the system with the neural algorithm. The physics of the system can take the form of, for example, a parameterized input to the algorithm (e.g., the noise level in a denoising system~\cite{gharbi2016deep}) or a sub-module in the algorithm architecture (e.g., the spatial transformer~\cite{jaderberg2015spatial}). In a camera array, the intrinsics and extrinsics are usually exploited. Using them to characterize an array have several advantages: 1) they are not likely to change once the cameras are encapsulated; 2) their derivatives like the epipolar geometry naturally build connections of sensor pixels; 3) they can be achieved by mature calibration techniques like \cite{zhang2000flexible} with efficiency. The epipolar transformer~\cite{he2020epipolar} is an example that leverages the epipolar geometry to estimate the human pose. Within the field of image fusion, the parallax-aware attention model ~\cite{wang2019learning} was introduced to derive a high-resolution image from two rectified low-resolution images. %
The model has been extended to process unrectified pairs~\cite{chen2021cross} and to solve general image restoration tasks from homogeneous views~\cite{yan2020disparity}. %
Our algorithm is also inspired by this architecture but focuses on general fusion tasks in the camera array.

Synthetic data is highly effective in training imaging systems with well-characterized forward models.  Array cameras, in particular, are easily simulated in the forward process. The forward model accounts for both optical components (e.g., lenses) and electronics (e.g., sensors) scene transformations. Datasets that include synthetic data can be semi-virtual, containing synthetic labels from real media of high quality such as the DND denoising benchmark~\cite{plotz2017benchmarking}, the Flickr1024 stereo super-resolution dataset~\cite{Flickr1024}, and the KITTI2012 multitask vision benchmark~\cite{Geiger2012CVPR}. On the other hand, datasets can be completely virtual from source to sensors, among those the MPI-Sintel Dataset~\cite{Butler:ECCV:2012} is one of the milestones that use CG software to generate data. It contains rich scenes and incredible labeling accuracy of the optical flow, depth, and segmentation, which are either not achievable or expensive to generate in real scenes. Its successors include  FlyingChairs~\cite{dosovitskiy2015flownet} and Scene Flow Datasets~\cite{mayer2016large}. As CG software keeps evolving, we see more fancy features being developed and integrated into handy packages, like BlenderProc~\cite{denninger2020blenderproc}, to create and render scenes that most resemble the real world. With the assistance of better synthetic data, we can expect networks of better performance to be easily deployed.

\section{Our Method}
The goal of PAT is to fuse data from cameras in an array. The fusion result reflects the viewpoint of one selected camera. The selected viewpoint is the alpha viewpoint $\alpha$ while others are alternative beta viewpoints $\beta_1 \sim \beta_m$, where $m$ is the number of alternative  viewpoints. To represent images, features, or parameters from a certain viewpoint, the viewpoint symbol is marked as the left superscript. 

\begin{figure*}[htbp]
\centering
\subfigure{
  \includegraphics[width=.9\linewidth]{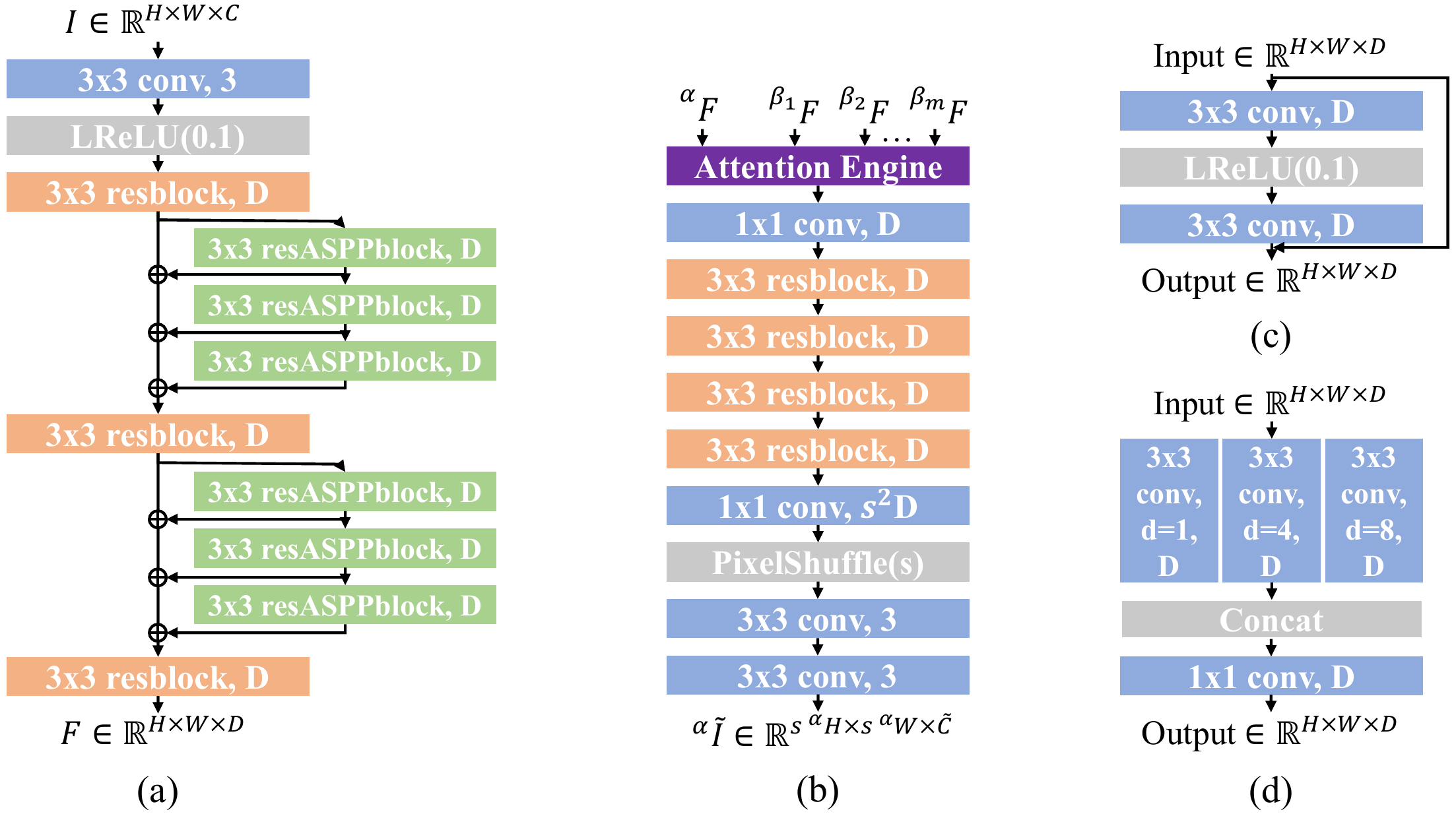}
}
\caption{The architecture of PAT. "$a\times a$ conv" (blue blocks) are convolutional layers with kernels of $a\times a$ size. The stride of convolutional layers is 1 and the padding is $(a-1)/2$. The dilation of convolutional layers $d$ is 1 unless specified. The depth dimension of convolutional layers and residual blocks (orange and green blocks) are notated after commas. The parameter each gray block may take is inside the parenthesis. (a) Image representation module. It takes sensor image $I$ and produces the associate feature $F$. $\oplus$ denotes addition. (b) The attention engine and post-fusion module. ${}^{\alpha}F$ is the feature of ${}^{\alpha}I$ from alpha viewpoint and ${}^{\beta}F$s are features of ${}^{\beta}I$s from alternative viewpoints. $s$ is the upscaling factor. The output is the fusion result ${}^{\alpha}\tilde{I}$. (c) The architecture of "$3\times 3 $ resblock". (d) The architecture of "$3\times 3$ resASPPblock".}
\label{fig:transformer-architecture}
\end{figure*}

The architecture of PAT is illustrated in Figure \ref{fig:transformer-architecture}. The workflow of PAT is 1) each sensor image goes through Figure \ref{fig:transformer-architecture}(a) to generate its corresponding feature; 2) features are fed into Figure \ref{fig:transformer-architecture}(b) to generate the final output. 
\subsection{Attention Engine}

The attention engine densely aligns the features with regard to the input physical receptive fields. The attention engine starts from image representations (features) of sensor images. As images may differ in resolution, color spaces, focal planes, etc., proper representations are learned to facilitate correspondence matching. We adopt the Residual ASPP Module~\cite{wang2019learning} to fulfill this task, which demonstrates effectiveness to generate multiscale features. For each camera, the feature produced by the image representation module shares the spatial dimensions with its sensor image, thus the position of each pixel is inherently encoded to the indices of voxels. Note here we do not require the resolution of all the cameras in the array to be the same, thus features may have diverse spatial dimensions but share the third dimension $D$. The representation modules of different input frames share the weights.

\begin{figure}[!ht]
\centering
\subfigure{
  \includegraphics[width=.3\linewidth]{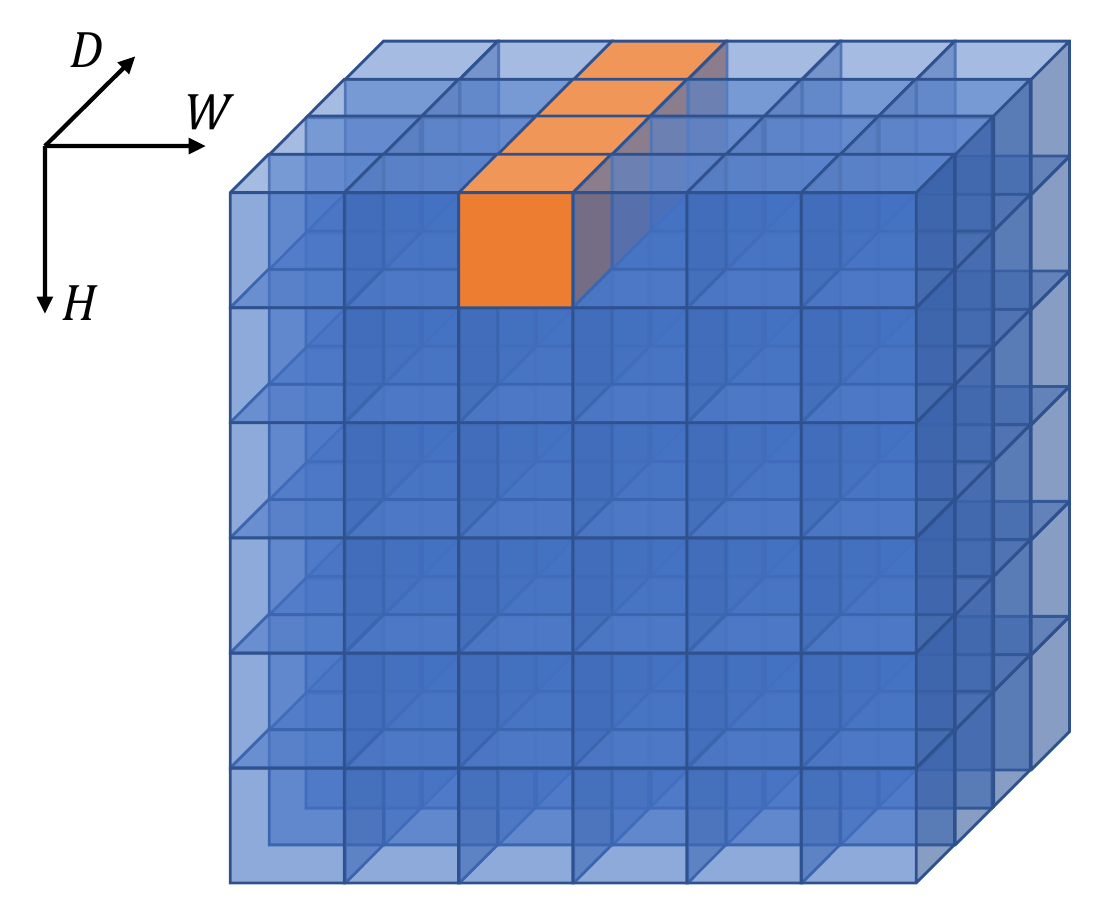}
}
\caption{The diagram of a feature (blue cube) and one of its voxels (orange stick).}
\label{fig:voxel}
\end{figure}

\begin{figure*}[htbp]
\centering
\subfigure{
  \includegraphics[width=.7\linewidth]{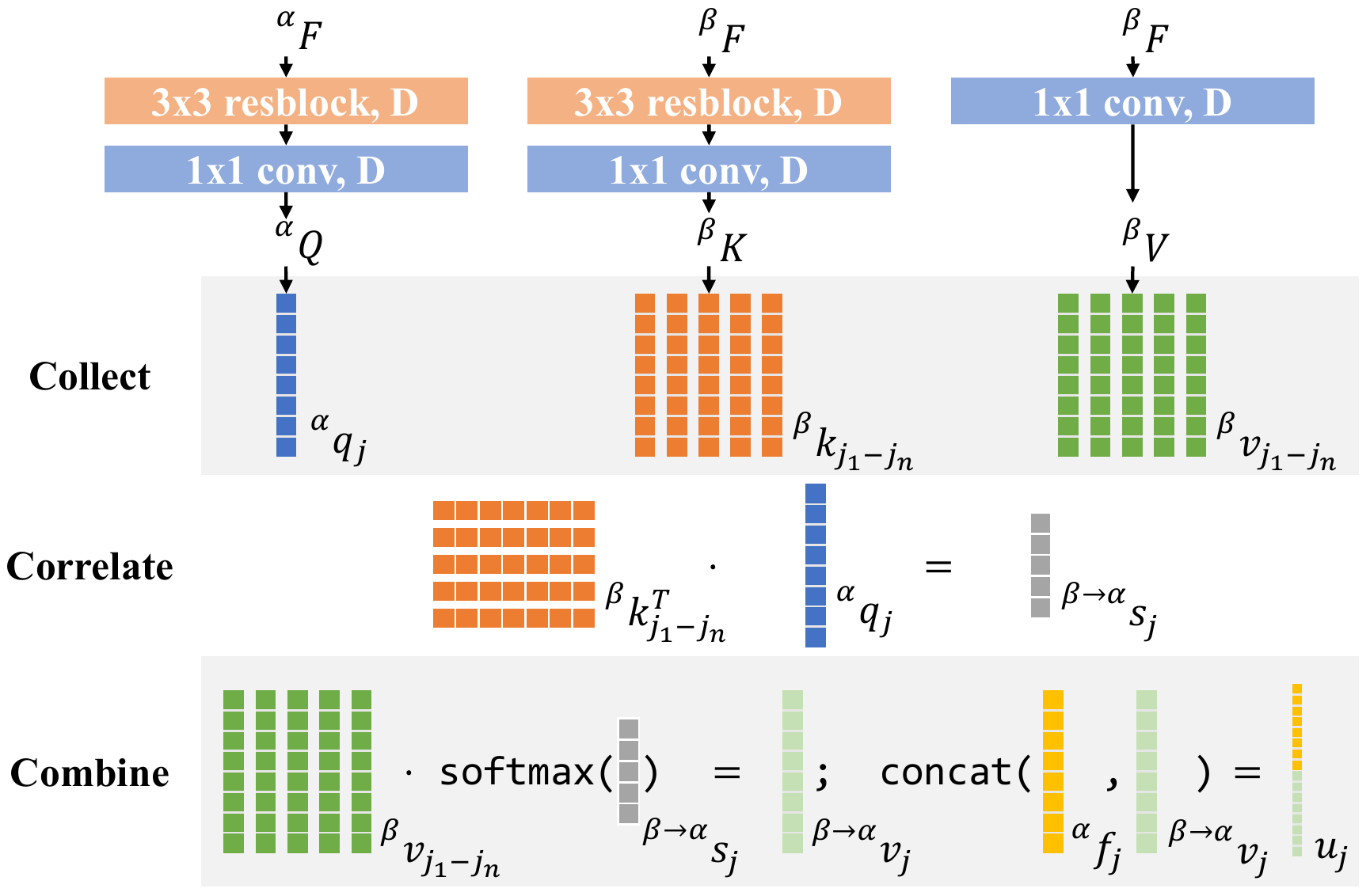}
}
\caption{Workflow of the attention engine and $\mathbf{C^3}$ operations in a dual-camera array. We use color bars to represent vectors and matrices. "$\cdot$" is the symbol of matrix multiplication.}
\label{fig:c-cube-operation}
\end{figure*}

We apply dot-product attention to compare and transfer the alternative features. From the feature ${}^{\alpha}F \in \mathbb{R}^{{}^{\alpha}H\times {}^{\alpha}W\times D}$ of the alpha camera, we produce a query feature ${}^{\alpha}Q \in \mathbb{R}^{{}^{\alpha}H\times {}^{\alpha}W\times D}$ through a residual block and a convolutional layer. Similarly, each alternative feature ${}^{\beta}F \in \mathbb{R}^{{}^{\beta}H\times {}^{\beta}W\times D}$ produces a key feature ${}^{\beta}K \in \mathbb{R}^{{}^{\beta}H\times {}^{\beta}W\times D}$ and a value feature ${}^{\beta}V \in \mathbb{R}^{{}^{\beta}H\times {}^{\beta}W\times D}$. We perform $\mathbf{C^3}$ operations: \textbf{Collect}, \textbf{Correlate}, and \textbf{Combine} to generate the feature output.
For simplicity, we use a system with two viewpoints to illustrate $\mathbf{C^3}$ operations, as shown in Figure \ref{fig:c-cube-operation}. 
\textbf{Collect}: ${}^{\alpha}q_j \in \mathbb{R}^D$ is $j$-th voxel in ${}^{\alpha}Q$, where a "voxel" denotes a $D$-length vector along the feature dimension, as illustrated in Figure \ref{fig:voxel}. The range of $j$ is from 1 to ${}^{\alpha}H{}^{\alpha}W$.  Voxels $\{ {}^{\beta}k_{j_1}, {}^{\beta}k_{j_2}, {}^{\beta}k_{j_3}, \dots, {}^{\beta}k_{j_n}\}, {}^{\beta}k \in \mathbb{R}^D$ in ${}^{\beta}K$  and voxels $\{ {}^{\beta}v_{j_1}, {}^{\beta}v_{j_2}, {}^{\beta}v_{j_3}, \dots, {}^{\beta}v_{j_n}\}, {}^{\beta}v \in \mathbb{R}^D$ are selected according to the receptive field of ${}^{\alpha}q_j$, where $j_n$ depends on $j$, ranging from 1 to ${}^{\beta}H{}^{\beta}W$. %

\textbf{Correlate}: we calculate the score ${}^{\beta\rightarrow \alpha}s_j$ between $ {}^{\alpha}q_j $ and ${}^{\beta}k$s to find the correspondence. ${}^{\beta\rightarrow \alpha}s_j$ is equal to the dot product of $\left[ {}^{\beta}k_{j_1}^T; {}^{\beta}k_{j_2}^T; {}^{\beta}k_{j_3}^T; \dots; {}^{\beta}k_{j_n}^T \right]$ and $ {}^{\alpha}q_j $. The $i$-th entry of ${}^{\beta\rightarrow \alpha}s_j$ is the score of $j_i$-th voxel. 

\textbf{Combine}: we combine voxels $\{ {}^{\beta}v_{j_1}, {}^{\beta}v_{j_2}, {}^{\beta}v_{j_3}, \dots, {}^{\beta}v_{j_n}\}$ with regard to $ {}^{\beta\rightarrow \alpha}s_j$ by calculating the dot product of $\left[  {}^{\beta}v_{j_1}, {}^{\beta}v_{j_2}, {}^{\beta}v_{j_3}, \dots, {}^{\beta}v_{j_n}\right]$ and $ \texttt{softmax}({}^{\beta\rightarrow \alpha}s_j)$, where \texttt{softmax}($\cdot$) is the softmax function. We denote the combined voxel as $ {}^{\beta\rightarrow \alpha}v_j$. The concatenation of ${}^{\alpha}f_j$ and $ {}^{\beta\rightarrow \alpha}v_j $ is the $j$-th voxel $u_j$ of the output feature $U \in \mathbb{R}^{{}^{\alpha}H\times {}^{\alpha}W \times 2D}$. 

For more than one alternative viewpoints, the $j$-th output vector $u_j$ is equal to $\texttt{concat}({}^{\alpha}f_j, {}^{\beta_1\rightarrow \alpha}v_j, {}^{\beta_2\rightarrow \alpha}v_j, \dots, {}^{\beta_m\rightarrow \alpha}v_j)$, for \texttt{concat}($\cdot$) is the concatenate function. $\mathbf{C^3}$ operations are fully vectorized, thus deployment of the attention engine on trending deep learning platforms is of convenience. %

The output feature $U \in \mathbb{R}^{{}^{\alpha}H\times {}^{\alpha}W \times (m+1)D}$ goes through several convolutional layers and residual blocks to produce the image output.%

One may notice the attention engine reduces to PAM~\cite{wang2019learning} when $m = 1, {}^{\alpha}H = {}^{\beta}H, {}^{\alpha}W = {}^{\beta}W, C=3, D=64$, and each receptive field follows the epipolar geometry in a rectified stereo pair, except that the intercorrelated validation mask is not incorporated into $U$. In comparison, PAT can integrate other physical clues, like maximum disparity or homography-based approximation, to optimize computation. Moreover, PAT can process images of different characteristics from three or more cameras, where rectification cannot be performed. %

\subsection{Complexity Analysis}
It is essential to ensure the above operations is achievable and efficient with regard to time complexity. For $j$ is the voxel index of the output feature $U$, let us assume $L = \max_{j} |\{~^\beta k_{j_1}, ~^\beta k_{j_2}, ..., ~^\beta k_{j_n}\}|$, where $|\cdot|$ returns the size of the set. %
The complexity of \textbf{Collect} is $O(L)$, of \textbf{Correlate} is $O(D\times L)$, and of \textbf{Combine} is $O(D\times L)$. Hence overall to compute the entire output feature for $m$ alternative viewpoints we have the complexity $O(m \times H\times W\times D\times L)$. 

For example, we assume we know the intrinsics and extrinsics in a dual-camera array, where $m = 1$ and the resolution of the cameras is $H\times W$. We can specify the physical receptive field in the attention engine to be the indices of beta feature voxels along the epipolar line of each alpha feature voxel, assuming the feature representations of images also follow the epipolar geometry in the spatial dimensions. If the baseline in a dual-camera system intersects both image planes at infinity, the epipolar line is always horizontal, i.e., $L = W$, bringing the total time complexity to $O(H\times W\times D\times W)$, which agrees with the complexity of PAM in~\cite{wang2019learning}. In systems of which the layout does not meet the condition above, $L$ is still linear to $H+W$, thus no extra time expense is introduced under the big $O$ notation. Furthermore, we can incorporate homography-based approximation, where we roughly locate associate voxels of each alpha voxel via perspective transformation. Knowing the depth range of interest in a scene, we can set the maximum pixel displacement $l$ between the rough estimates and exact correspondences to truncate the epipolar lines. Thus the complexity can be further reduced to $O(H\times W\times D\times l)$, as typically $l \ll W$.

Here we can see another merit the attention engine carries in terms of time complexity. We can chop the alpha feature into patches with spatial resolution $H_p \times W_p$. The attention engine infers on those patches in parallel, bringing the time complexity down to $O(m \times H_p \times W_p \times D \times L)$. %

\subsection{Data synthesis}
As we stated in the introduction, %
the pipeline of data simulation is fully automatic. We use the Python API of Blender to scale up the generation of scenes. Blender provides a variety of meshes, from which we select several representative meshes including 'plane', 'cube', and 'uv-sphere', and enrich the database by perturbing the surface to create diverse ridges and valleys. We can specify the dimensions, locations, and rotations of the meshes to diversify their distribution in a scene. Upon the creation of each shape, we can attach the 'material' attribute to customize its interaction with the light source. There are dozens of knobs to adjust the base color, diffusion, or specularity; apart from those, we can apply vectorized textures, e.g., brick texture and checkerboard texture, to add varieties to color distribution on the mesh. Occlusion and shadows are naturally introduced while stacking up the meshes. %

\begin{figure}[htbp]
\centering
\subfigure{
  \includegraphics[width=.45\linewidth]{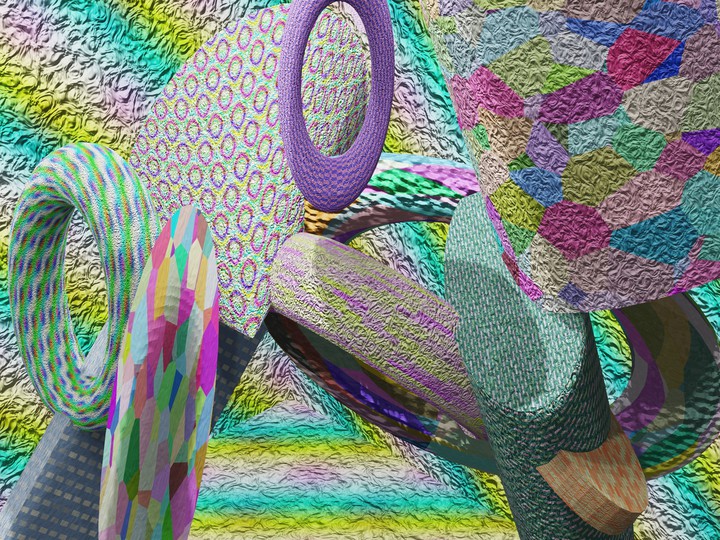}
}
\subfigure{
  \includegraphics[width=.45\linewidth]{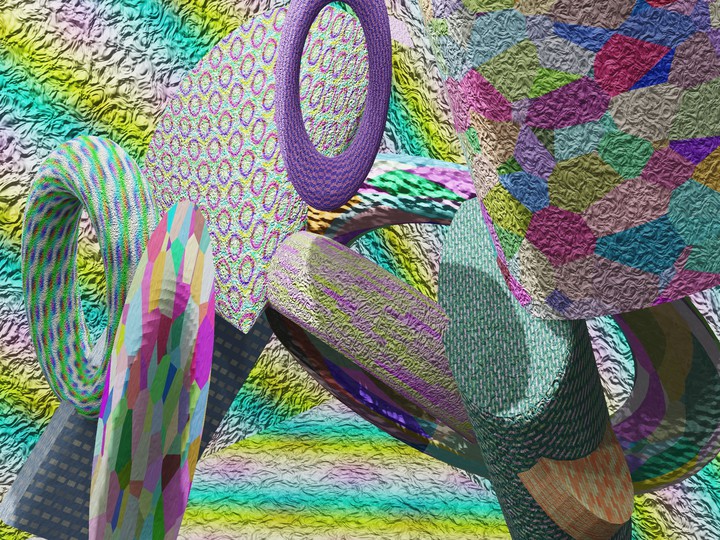}
}
\caption{An example of the views of a dual camera system. The resolution of the images is $2048\times 1536$.}
\label{fig:data-sim-example}
\end{figure}

Blender provides camera objects to render the scene. Just as in real cameras, parameters like the focal length, sensor size, pixel pitch, and resolution can be easily set. If the 'Depth of field' feature is on, parameters like the focal plane and F-stop allow realistic modeling of the defocus blur. Blender allows common picture formats as outputs, including lossy JPEG, lossless PNG, or even RAW with full float accuracy. The color space of the output can be BW, RGB, or RGBA. Figure \ref{fig:data-sim-example} shows an example of rendered views of a dual-camera system. We can see rich features, colors, and interactions of the objects in the frames, and also parallax between two frames.

We also implement other functions, among which we would like to emphasize the function of animation generation. We can assign random trajectories and transformations to mashes, and stream the data with regard to a given frame rate. It can benefit array camera research on temporal connections.

\subsection{Implementation Details}
The following details are shared by PATs for the experiments. The unique settings for each application are specified in the next section.

\paragraph{Dataset} We rendered 900 scenes of the resolution $2048\times1536$ to two cameras using the EEVEE render engine in Blender 2.92. 800 scenes were for the training and 100 scenes for the validation. One of the virtual cameras was selected to have the alpha viewpoint and the other had the beta viewpoint.%
The objects in the scene distribute within a 20-meter range. Rendered frames were in RGB color. We selected 49 patches of the resolution $384\times 128$ across each scene, and then cubically downsampled the patches to $96\times32$. Each sample in the dataset has a pair of patches, where the patch from the alpha view is regarded as the ground truth. The degraded inputs, instead, were generated from the patch from the alpha view (alpha patch) and beta view (beta patch) while training via the forward model with regard to the array setting. We exported the extrinsics and intrinsics of two cameras and constructed the receptive fields according to the epipolar geometry, i.e., a dense map from each voxel index $j$ in the alpha view to the associate voxel indices $j_1 \sim j_n$ in the beta views. $n$ for all $j$ was set to 96 in our dataset. The physical receptive fields for each sample were stored as arrays along with two patches. 

\paragraph{Training}
PATs were trained on the NVIDIA Tesla V100S GPU. Hyperparameters below were shared by the experimental systems:
\begin{center}
\begin{tabular}{ l l l l l}
 $\mathbf{D}$ & 64 &   & \textbf{Epoch} & 80\\ 
 $\mathbf{s}$ & 1 &   & \textbf{Loss} & Mean Square Error\\
 $\mathbf{\tilde{C}}$ & 3 &   & \textbf{Optimizer} & Adam~\cite{kingma2014adam}\\
  \multicolumn{2}{l}{\textbf{Learning Rate}}  & \multicolumn{3}{l}{0.0002, decays by half per 30 epochs} 
\end{tabular}
\end{center}
\noindent The parameters that were specific to the application are clarified in the following subsections. The model with the best peak signal-to-noise-ratio (PSNR) performance on the validation set was selected for inference.

\paragraph{Inference} The intrinsics and extrinsics of the camera array were calibrated through MATLAB Stereo Vision Toolbox. We combined epipolar geometry and homography-based approximation to construct the physical receptive fields. The max displacement $l$ is set to 80.

\section{Experiments}
Here we demonstrate four experimental systems with diverse sampling designs and PAT processing for image fusion, following the order mentioned in the introduction. We have posted the training and evaluation code for all of these experiments online~\cite{code}.  %

\subsection{Wide Field - Narrow Field System}
\label{sec:wide-narrow}
\begin{figure*}[htbp]
\centering
\subfigure[Narrow field camera view]{
  \includegraphics[width=.3\linewidth]{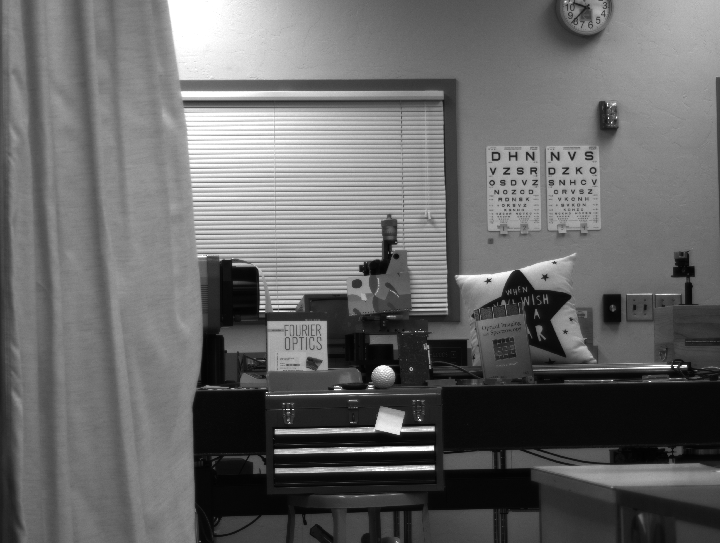}
}
\subfigure[Wide field camera view]{
  \includegraphics[width=.3\linewidth]{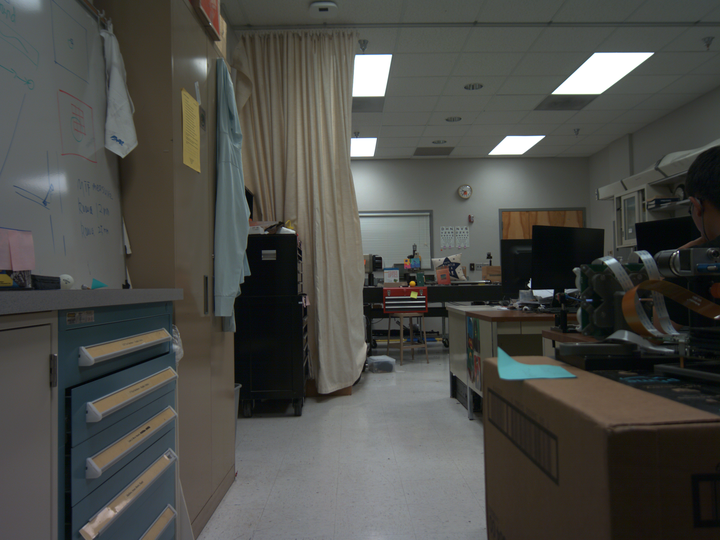}
}
\subfigure[Wide field camera view (patch)]{
  \includegraphics[width=.32\linewidth]{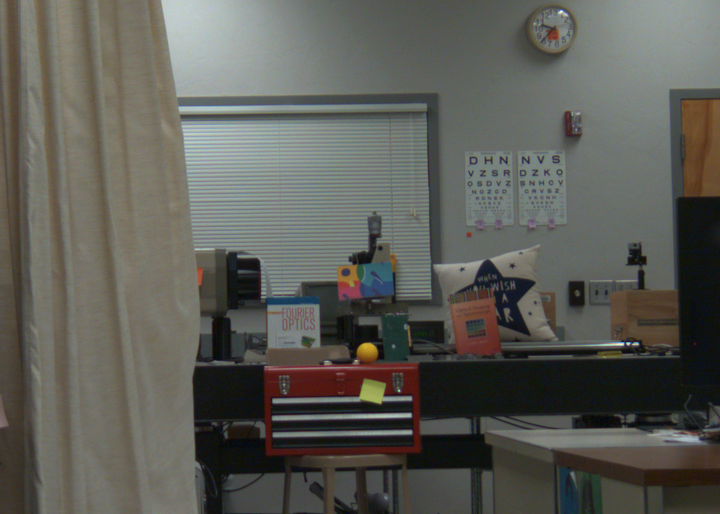}
}
\caption{The views of the wide field - narrow field array. The pixel count of (a) is around $10\times$ the pixel count of (c). Images are rescaled for display purposes.}
\label{fig:dual-vision-platform}
\end{figure*}

It is observed that chroma can be substantially compressed compared to luminance before the decompression error is perceived by humans. Inspired by that, we demonstrate a wide field color - narrow field monochrome system that compressed the color of the narrow field of view (FoV) by up to $40\times$.
The configurations of the array were:%
\begin{center}
\begin{tabular}{l l}
 \multicolumn{2}{l}{Narrow field camera}\\
 \textbf{Body} & Allied Vision Alvium 1800 U-1240m\\  
 \textbf{Sensor} & CMOS Monochrome\\ 
 \textbf{Lens} & 25 mm TECHSPEC HR Series \\ 
 \textbf{Resolution}& $4024\times 3036$ \\ 
 \\
 \multicolumn{2}{l}{Wide field camera}\\
 \textbf{Body} & iDS UI-3590LE-C-HQ\\  
 \textbf{Sensor} & CMOS Color\\ 
 \textbf{Lens} & 5 mm Kowa LM5JCM \\ 
 \textbf{Resolution}& $4912\times 3684$
\end{tabular}
\end{center}
\noindent The focal plane of the wide field camera was set to its hyperfocal distance. The narrow field camera focused on the black optical table around 7 meters away. 

Figure \ref{fig:dual-vision-platform} shows the camera views in an example scene. Considering the color filters on the wide field camera and $10\times$ resolution gap in the narrow field, the red and blue raw signals were subsampled by $10\times4 = 40\times$ and the green raw signals were subsampled by $10\times2 = 20\times$ compared to the luminance. %

PAT acted as a color decompressor on this system that upsampled the colors in the narrow field. %
PAT was trained with two inputs. We converted the alpha patch to grayscale as one input and had the beta patch unchanged as the other input. To model possible resolution gaps and defocus blur, we augmented the training data by 1) adding box blur to the alpha input; 2) adding box blur to the beta input; 3) $2\times$ bicubically downsampling the beta input; 4) combining 2) and 3). These augmentation techniques were selected at random with equivalent probabilities during training and validation. The batch size of training was set to 32 and $C$ was set to 3. In the training and inference phases, the alpha input was repeated along the feature dimension three times and the beta input was bicubically upsampled to its original dimensions if it had been downsampled. 

Before implementing the trained PAT on the system, we evaluated the algorithm on Flickr1024~\cite{Flickr1024} and KITTI2012~\cite{Geiger2012CVPR} (20 frames) test sets. For each testing sample, we used whole frames instead of patches to generate inputs. The alpha frame was converted to grayscale as the alpha input and the beta frame was $2\times$ or $4\times$ bicubically downsampled as the beta input. Based on the characteristics of the test sets, the physical receptive fields indicated truncated horizontal epipolar lines of the length 120 divided by the downsampling rate. In Table \ref{tab:pat-stereo-stats}, we listed average PSNR and SSIM~\cite{wang2004image} scores between 1) the ground truth alpha frames and the grayscale alpha inputs in the "Alpha Input" column; 2) the beta frames and the beta inputs bicubically-upsampled to the original size in the "Beta Input" column; 3) the ground truth alpha frames and the fused results of PAT in the "Fusion" column. We can see PAT improved the test system by maintaining the structures of the alpha input and improving the color upsampling results compared to the beta input solely.
\begin{table}[!ht]
\caption{Comparison between inputs and fused results of PAT (monochrome - color inputs)}
\label{tab:pat-stereo-stats}
\begin{center}
{\begin{tabular}{ |c|c|c|c|c| } 
 \hline
 Dataset & Scale & Alpha Input & Beta Input & Fusion \\ 
   \hline
 \multirow{2}{*}{Flickr2014}& x2 & \multirow{2}{*}{21.42/0.8800} & 24.95/0.8161 & 27.26/0.8992\\ 
  & x4 &  & 21.84/0.6265 & 25.85/0.8840\\ 
  \hline
 \multirow{2}{*}{KITTI2012} & x2 & \multirow{2}{*}{26.40/0.9178} & 28.48/0.8845 & 29.55/0.9097\\ 
  & x4 &  & 24.56/0.7376 & 28.40/0.8957\\ 
 \hline
\end{tabular}}
\end{center}
\end{table}

\begin{figure*}[htbp]
\centering
\subfigure{
  \includegraphics[width=.9\linewidth]{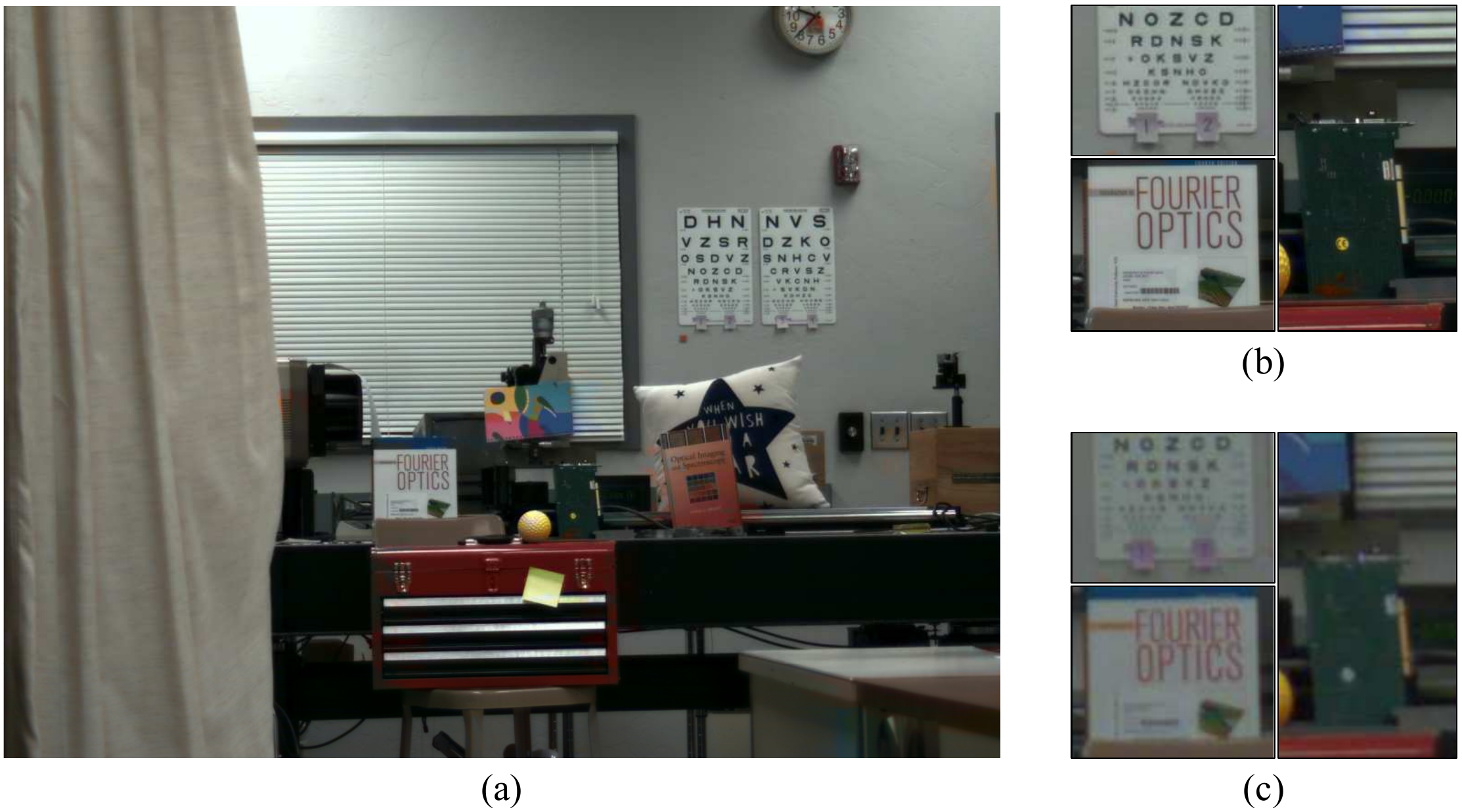}
}
\caption{(a) is the fused frame on the wide field - narrow field array. (b) and (c) are associate details of the fused frame and the view of the wide field camera, respectively.}
\label{fig:dual-vision-result}
\end{figure*}

\begin{figure*}[htbp]
\centering
\subfigure[]{
  \includegraphics[width=.45\linewidth]{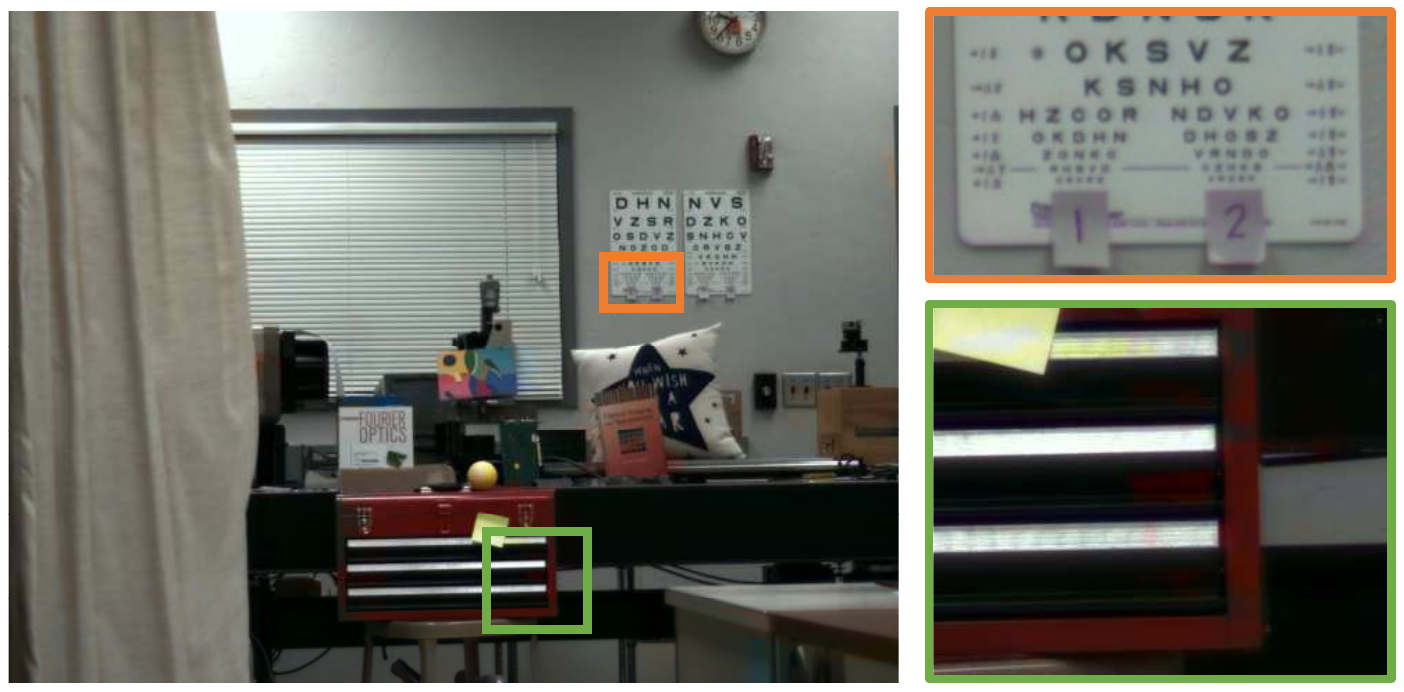}
}
\subfigure[]{
  \includegraphics[width=.45\linewidth]{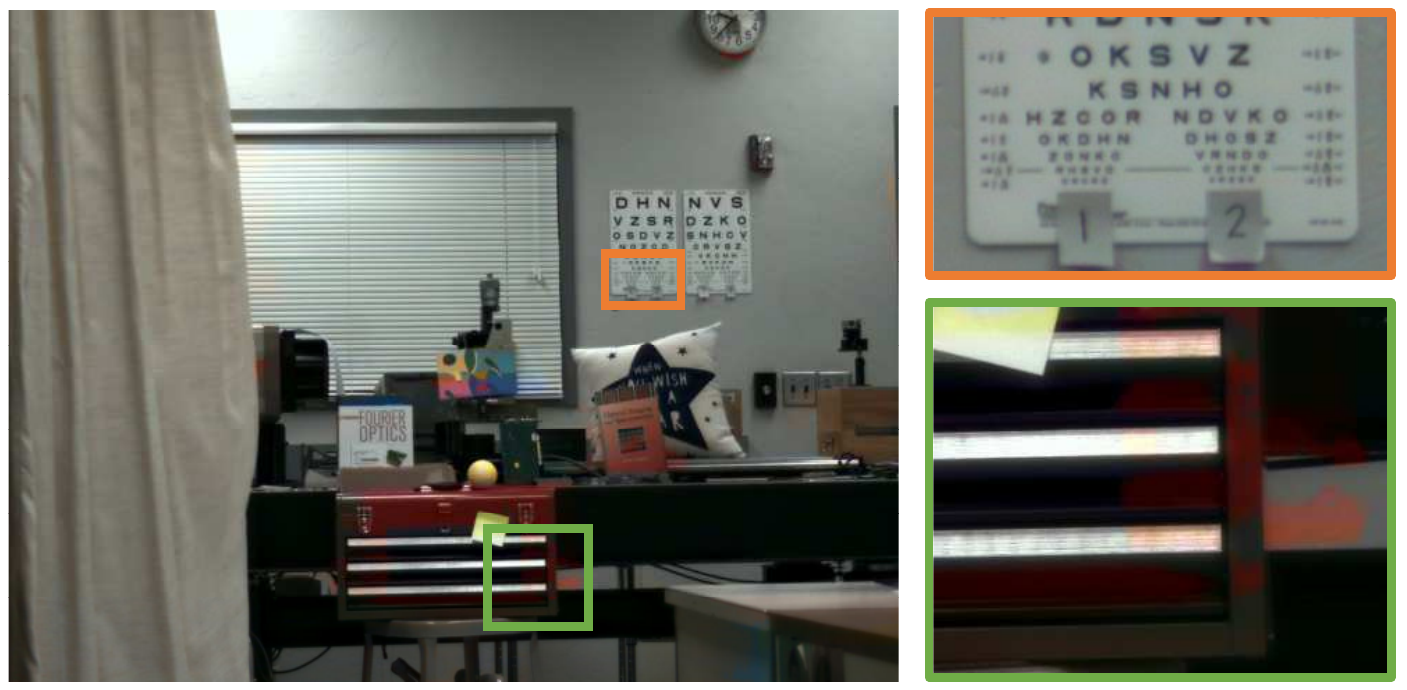}
}
\caption{The fused frame with different receptive fields. (a) is generated with the calibrated receptive fields that accurately reflect the physics of the system. (b) is generated with the receptive fields assuming the inputs are rectified.}
\label{fig:dual-vision-horizontal-result}
\end{figure*}

We assigned the alpha viewpoint to the narrow field camera while inferencing. The result is shown in Figure \ref{fig:dual-vision-result}. In comparison, colors were upsampled by up to $40\times$ to the narrow view without scarifying the sampling rate of the luminance. Although the color bleeding artifacts caused by a large upsampling rate can be observed in certain regions, we reduced the artifacts to the minimum by providing accurate physical information to the system. %
As illustrated in Figure \ref{fig:dual-vision-horizontal-result}, the correct receptive field yielded the result in Figure \ref{fig:dual-vision-horizontal-result}(a) with correct colors (pink stickers in the orange window) and less artifacts (storage box in the green window).

\subsection{Visible - Near Infrared Systems}
As a result of reduced atmospheric scatter and absorption, near infrared cameras achieve higher contrast in landscape photography. However, IR signals are typically recorded as monochromatic data, thus are not visual friendly. Here we show PAT acted as a visualization tool to fuse color and NIR views while retaining the texture of remote objects on  visible - NIR camera arrays. We used the data from two visible-NIR arrays, one was from a public database PittsStereo-RGBNIR~\cite{zhi2018deep} and the other was built by us\footnote{Thanks to Emily Chau in Infrared Imaging Group at the University of Arizona for the array setup and data collection.}. The configurations of the camera array from the online dataset are available in~\cite{zhi2018deep}. We used rectified images of the resolution $582\times 429$ from the database. Our visible-NIR system was composed of two 35 mm EO-4010 cameras, one with a color filter and the other with a NIR filter. The resolution of both cameras was $2048\times 2048$.

We applied the pretrained PAT from the wide field - narrow field system to this fusion task to highlight the ability of domain adaptation of our algorithm. The attention engine of PAT operates on the features, thus is robust to the data that differs in appearance, brightness, etc. 

\begin{figure*}[htbp]
\centering
\subfigure[NIR frame]{
  \includegraphics[width=.3\linewidth]{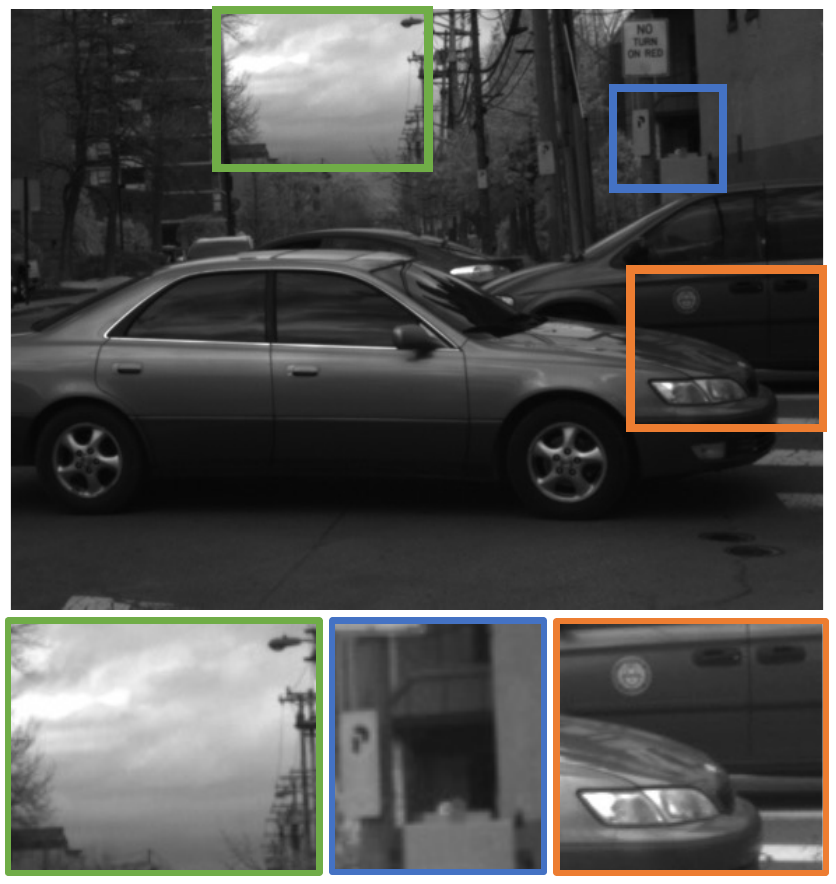}
}
\subfigure[Visible frame]{
  \includegraphics[width=.3\linewidth]{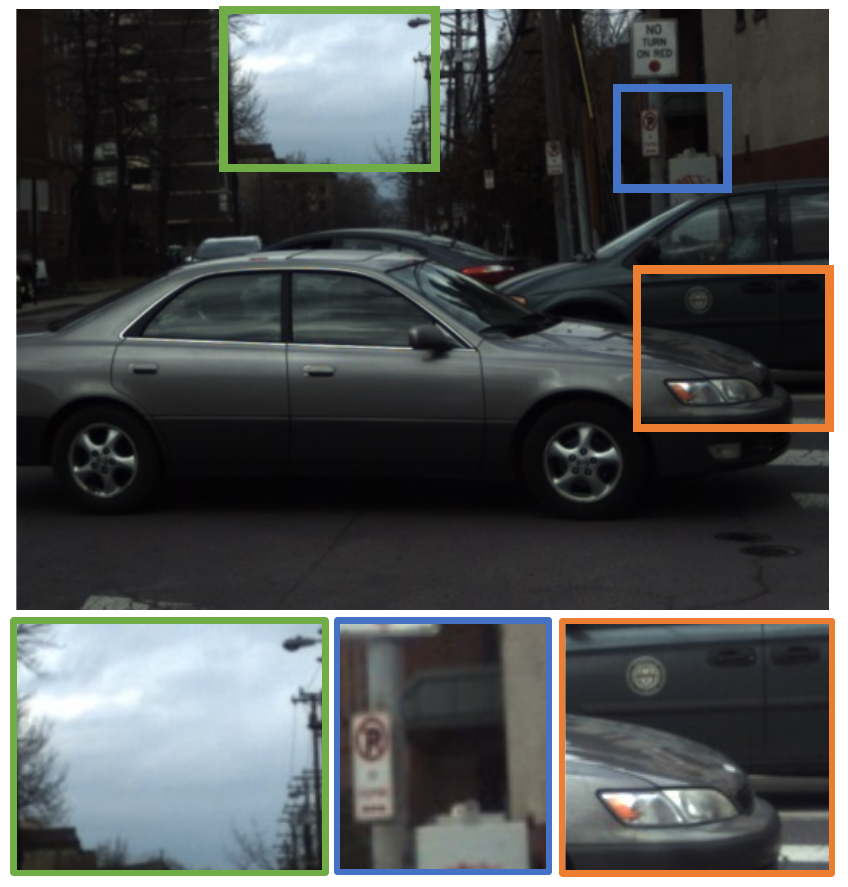}
}
\subfigure[The fused frame]{
  \includegraphics[width=.3\linewidth]{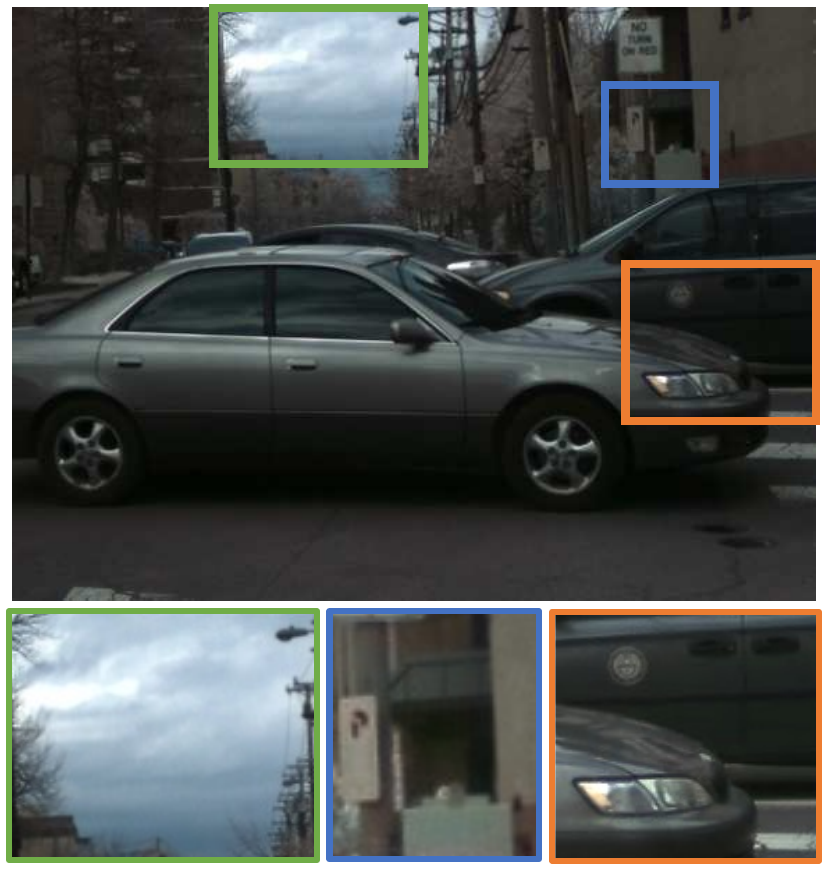}
}
\caption{Data from PittsStereo-RGBNIR dataset~\cite{zhi2018deep} and the fused result. The orange, blue, and green windows contain the details in the scene from near to far. The brightness of details is adjusted to enhance contrast.}
\label{fig:rgb-nir}
\end{figure*}

\begin{figure*}[htbp]
\centering
\subfigure[NIR frame]{
  \includegraphics[width=.3\linewidth]{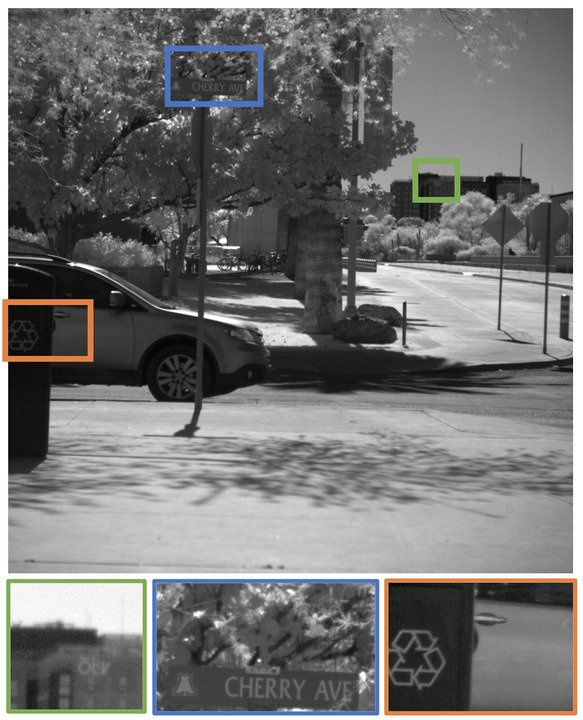}
}
\subfigure[Visible frame]{
  \includegraphics[width=.3\linewidth]{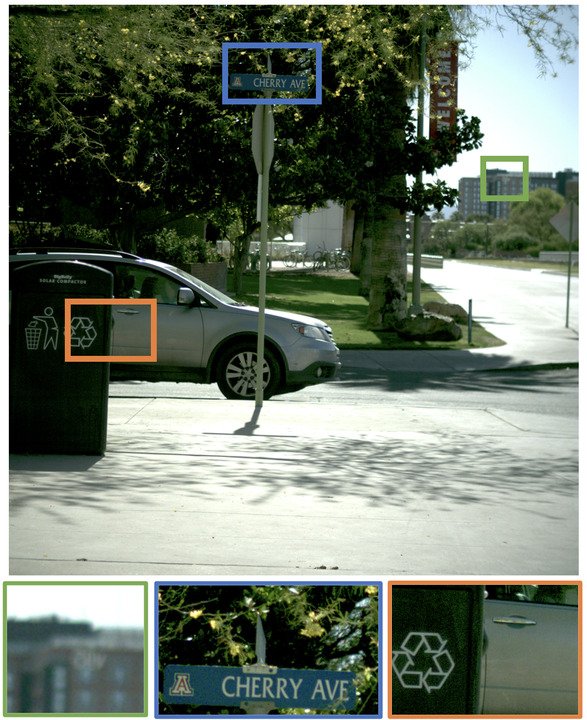}
}
\subfigure[The fused frame]{
  \includegraphics[width=.3\linewidth]{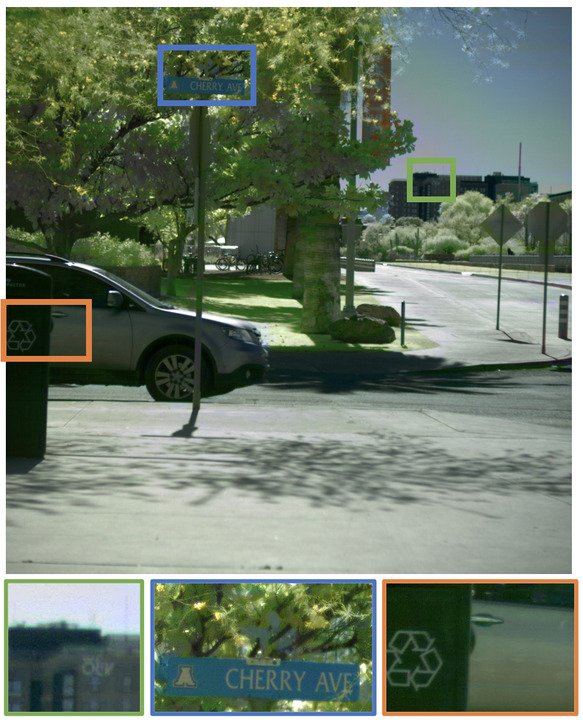}
}
\caption{Data from our visible-NIR camera array and the fused result. The orange, blue and green windows contain the details in the scene from near to far. The brightness of details is adjusted to enhance contrast.}
\label{fig:rgb-nir-driggerslab}
\end{figure*}

The alpha viewpoint was assigned to the NIR camera while inferencing. Figure \ref{fig:rgb-nir} and \ref{fig:rgb-nir-driggerslab} demonstrate the fusion results with zoomed-in details on the given data. The color was well transferred to the fusion results in the presence of complicated occlusion and parallax. Also different appearances of distant objects in the visible and NIR frames were fused nicely, as demonstrated in the green boxes.

\subsection{Short Exposure - Long Exposure System}
\begin{figure*}[htbp]
\centering
\subfigure[unfiltered camera view, 150 units exposure time]{
  \includegraphics[width=.45\linewidth]{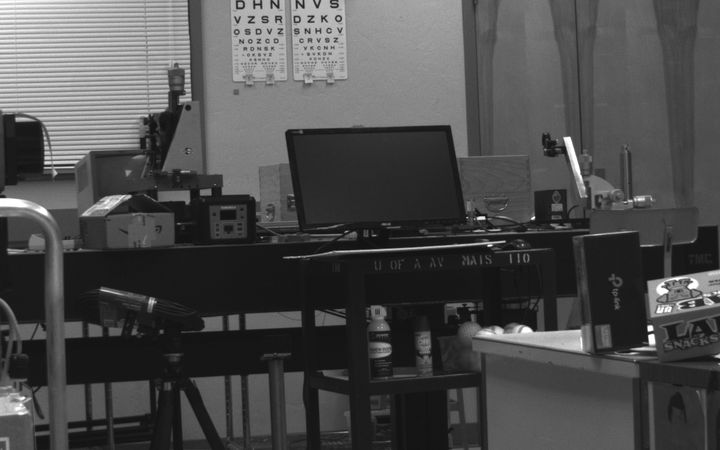}
}
\subfigure[450 nm camera view, 800 units exposure time ]{
  \includegraphics[width=.45\linewidth]{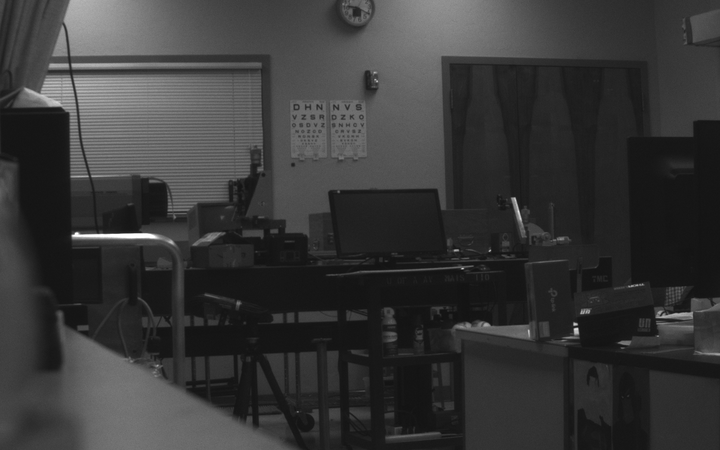}
}
\subfigure[550 nm camera view, 300 units exposure time]{
  \includegraphics[width=.45\linewidth]{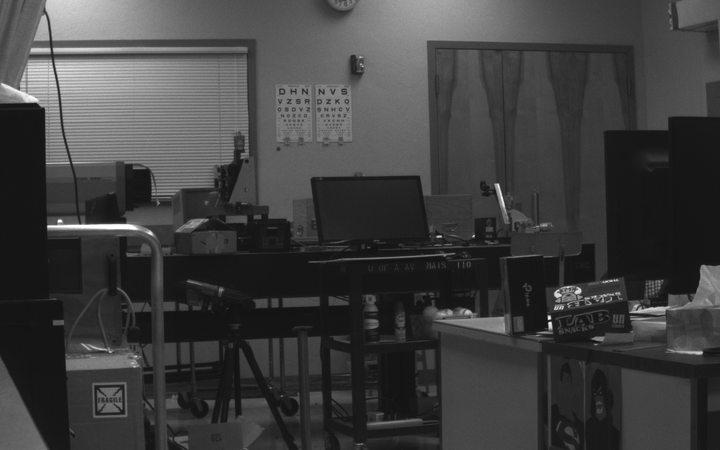}
}
\subfigure[600 nm camera view, 500 units exposure time]{
  \includegraphics[width=.45\linewidth]{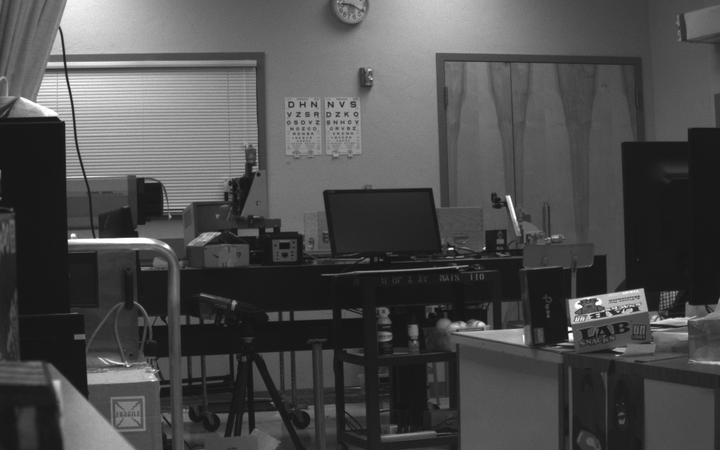}
}
\caption{The views of the short exposure - long exposure array under diverse exposures in relative scales. 1 unit is equal to approximately 12 micro seconds.}
\label{fig:2x2array-views}
\end{figure*}

For visible color imaging, multiaperture sampling allows independent exposure and focus control for each band. We demonstrate this capability using a $2\times2$ camera array based on the Arducam 1MP$\times 4$ Quadrascopic OV9281. %
The cameras were monochromatic and had $1280\times 800$ resolution. One camera with a 12 mm lens had no filter, while the others with 8 mm lenses were equipped with three filters. The central wavelengths of the filters were 450 nm, 550 nm, and 600 nm respectively. The filters shared 80 nm full width at half maximum. The exposure time of each camera was controlled independently to optimize the dynamic range of the signal. %
Figure \ref{fig:2x2array-views} shows the views of four cameras. Compared to uni-exposure systems like cameras with the Bayer filter, our system allowed up to $5\times$ differences in exposure, thus having higher overall throughput of spectral data. %

PAT was trained with 4 inputs. The alpha patch was converted to be grayscale as the alpha input. The red, green, and blue channels were unpacked from the beta patch as three alternative inputs. Note the color filters of our synthetic training data did not exactly resemble the filters we used with regard to the spectral curves. The batch size of training was set to 16 and $C$ was set to 1 as inputs were monochromatic.

Most of the test settings agreed with those in the wide field - narrow field system, except that the beta frame was unpacked into three frames of a single color channel and downsampled to generate three beta inputs. 
In Table \ref{tab:pat-multiframe-stats}, PSNR and SSIM scores in the "Beta Inputs" column were first averaged between three spectral bands of the beta frame and corresponding beta inputs, and then averaged across all beta frames. The meanings of other columns are the same as in Table \ref{tab:pat-stereo-stats}. Similarly we can also see PAT improved overall system performance.
\begin{table}[!ht]
\caption{Comparison between inputs and fused results of PAT (monochrome - spectral inputs)}
\label{tab:pat-multiframe-stats}
\begin{center}
{\begin{tabular}{ |c|c|c|c|c| } 
 \hline
 Dataset & Scale & Alpha Input & Beta Inputs & PAT \\ 
   \hline
 \multirow{2}{*}{Flickr2014}& x2 & \multirow{2}{*}{21.42/0.8800} & 24.95/0.8161 & 25.55/0.8714\\ 
  & x4 &  & 21.84/0.6265 & 24.00/0.8493\\ 
    \hline
 \multirow{2}{*}{KITTI2012} & x2 & \multirow{2}{*}{26.40/0.9178} & 28.48/0.8845 & 28.77/0.8955\\ 
  & x4 &  & 24.56/0.7376 & 28.13/0.8903\\ 
 \hline
\end{tabular}}
\end{center}
\end{table}

\begin{figure*}[htbp]
\centering
\subfigure[]{
  \includegraphics[width=.6\linewidth]{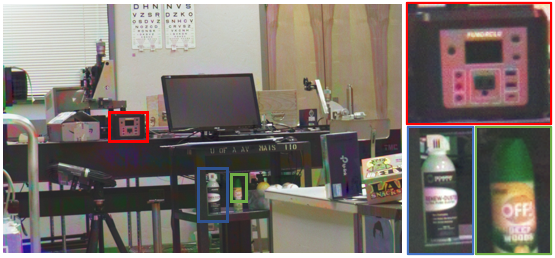}
}
\subfigure[]{
  \includegraphics[width=.156\linewidth]{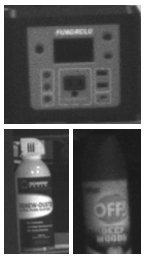}
}
\subfigure[]{
  \includegraphics[width=.16\linewidth]{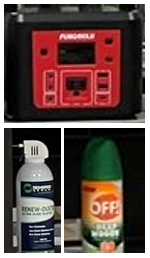}
}
\caption{(a) is the fused frame on the short exposure - long exposure array. Zoomed-in views on the side highlight objects of different spectral responses. (b) are the associate details of the original frame from the short exposure camera. Details in (c) are captured by a cellphone camera to provide the readers with color references. The color balance of the fused frame and the brightness of details are adjusted for visual purposes.}
\label{fig:2x2array-result-w-details}
\end{figure*}

\begin{figure*}[htbp]
\centering
\subfigure{
  \includegraphics[width=.45\linewidth]{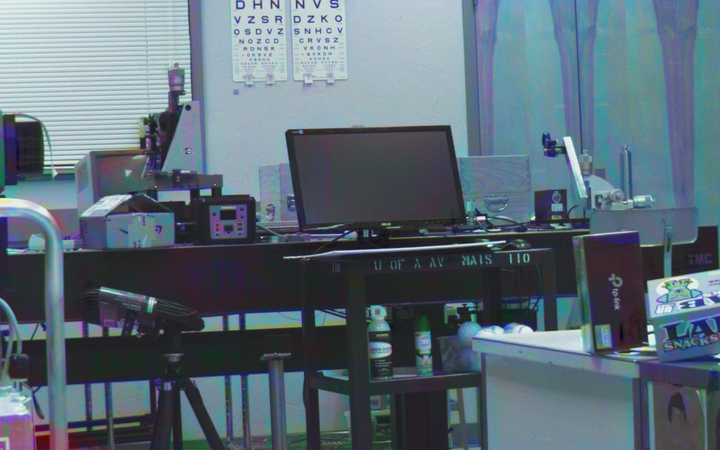}
}
\caption{The fused results of the permutated input sequence, where the 450 nm view and 650 nm view were switched. The color balance is adjusted for visual purposes.}
\label{fig:2x2array-permutate}
\end{figure*}

While inferencing, the alpha viewpoint was assigned to the camera without the filter. Figure \ref{fig:2x2array-result-w-details} shows the fused result. The result preserved the geometry of the alpha camera view and displayed the correct color, which indicates the algorithm effectively adapted to data with different filter functions. We can expect the result generated with the optimized spectral throughput to have a higher dynamic range. Note that PAT is physical-based rather than perception-based, therefore the network does not "guess" the color beyond physical clues. As shown in Figure \ref{fig:2x2array-permutate}, the color channels of the fused frame were permuted with regard to the way that the inputs were permuted.

\subsection{High Frame Rate - Low Frame Rate System}
Sensors with color filters sacrifice quantum efficiency compared to monochrome sensors, thus requiring a longer exposure time to achieve a comparable signal-to-noise ratio (SNR). This prevents standalone spectral cameras from achieving a higher frame rate. Here we demonstrate an imaging system that combines one high frame rate (HFR) monochrome camera with three low frame rate (LFR) spectral cameras is a better solution to sample the light field temporally. This system enables PAT to reconstruct the light field at a high frame rate.

\begin{figure*}[htbp]
\centering
\subfigure[HFR camera view]{
  \includegraphics[width=.45\linewidth]{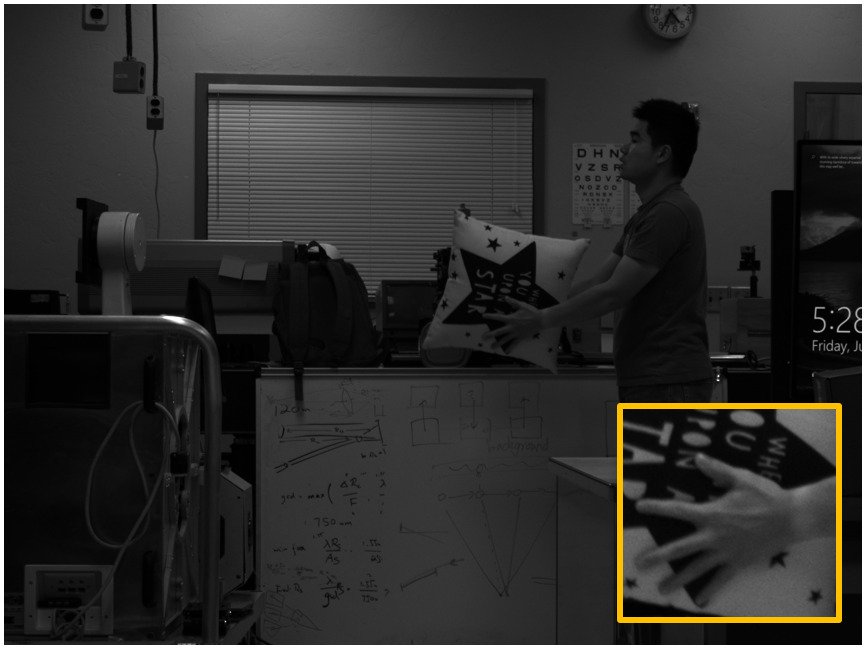}
}
\subfigure[450 nm LFR camera view]{
  \includegraphics[width=.45\linewidth]{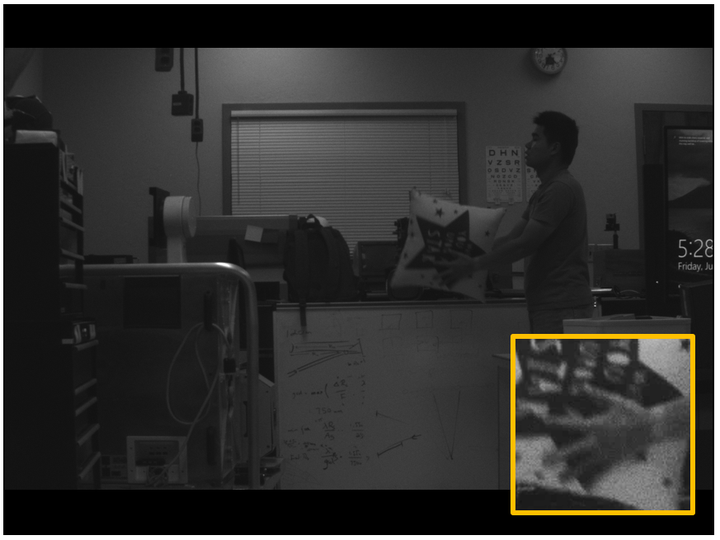}
}
\subfigure[550 nm LFR camera view]{
  \includegraphics[width=.45\linewidth]{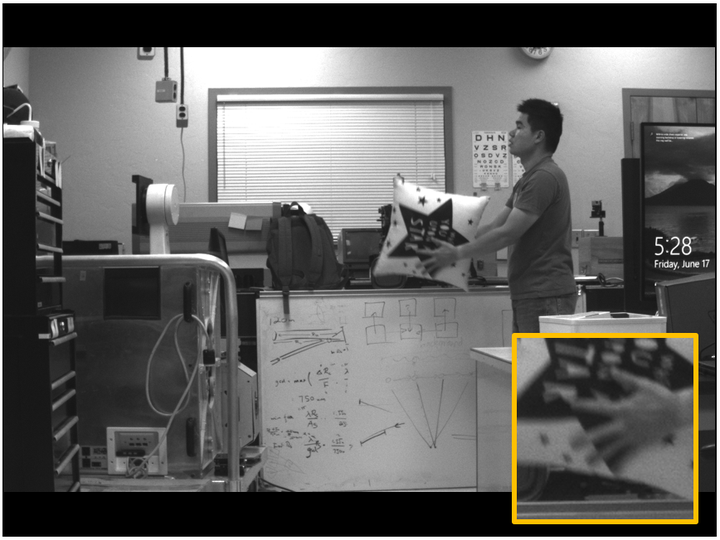}
}
\subfigure[600 nm LFR camera view]{
  \includegraphics[width=.45\linewidth]{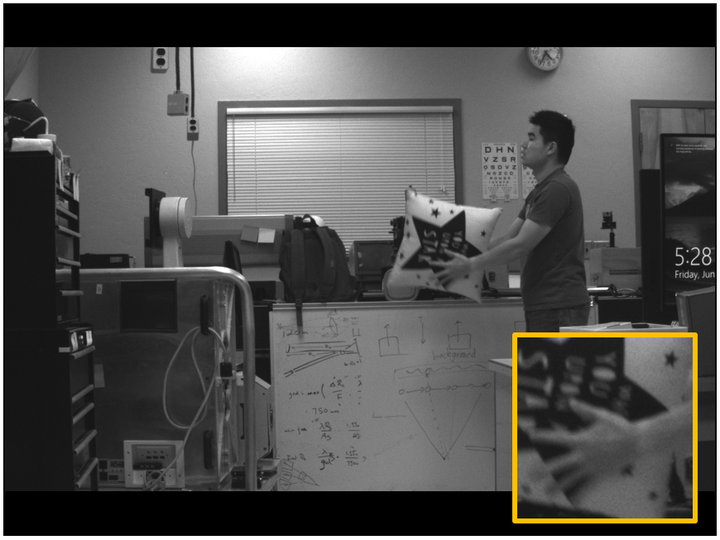}
}
\caption{The views of the HFR - LFR array. The exposure time of (a) was 4.3 ms. The exposure times of (b) - (d) were the same, around 12 ms. Since the HFR camera was not synchronized with LFR cameras, (a) was captured $\pm 2.15$ ms away from the moment that (b) - (d) were captured. The orange windows highlight the moving pillow. The brightness and contrast of the patches in the orange windows were adjusted for visual purposes.}
\label{fig:high-low-framerate-views}
\end{figure*}

We applied one Basler acA1440-220um camera with a 12 mm len as the HFR camera, which can reach 227 frames per second (fps) at the $1456\times1088$ resolution. Three Arducam cameras with 8 mm lenses and spectral filters in the short exposure - long exposure system were applied as the LFR cameras. The LFR cameras are synchronized, operating at 30 fps.
Figure \ref{fig:high-low-framerate-views} shows the views of four cameras in a scene where the moderate motion of the pillow occurred. We can see the LFR frames deteriorated in the region that has motions, while the associate region in the HFR frame remained sharp.

\begin{figure*}[htbp]
\centering
\subfigure{
  \includegraphics[width=.5\linewidth]{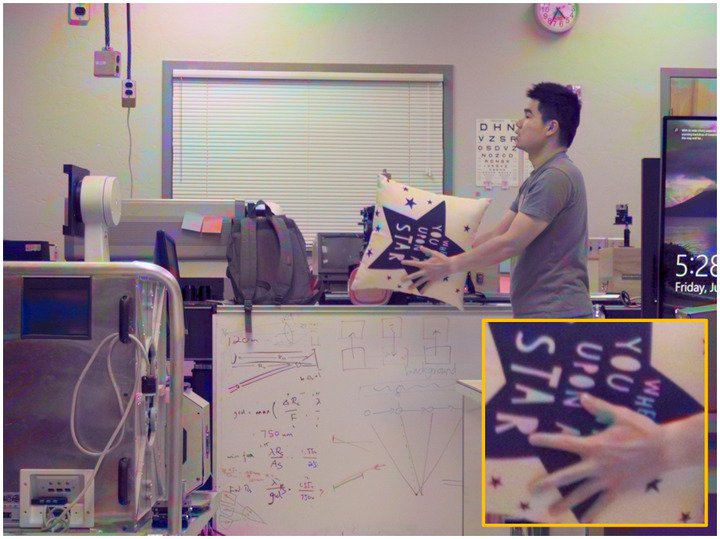}
 }
\caption{The fused results on the HFR - LFR array. The color balance is adjusted for display purposes.}
\label{fig:high-low-framerate-result}
\end{figure*}

The alpha and beta viewpoints were assigned to the HFR camera and LFR cameras, respectively. For one LFR frame captured at a certain moment, the HFR frames captured $\pm15$ ms from that moment correspond to that LFR frame. We assumed the epipolar constraint was valid in general between the LFR frame and associate HFR frames and built physical receptive fields accordingly. We applied the pretrained PAT from the short exposure-long exposure system while inferencing. Figure \ref{fig:high-low-framerate-result} shows the fused result, which fused the HFR camera view with colors from three spectral cameras. Because the epipolar geometry does not strictly hold for unsynchronized frames, slight color jittering of letters in the pillow was observed. However, the majority of colors of the pillow were effectively fused and the motion boundary was well preserved.

\begin{figure*}[htbp]
  \centering
  \subfigure[$0$ ms]{
  \includegraphics[width=.098\linewidth]{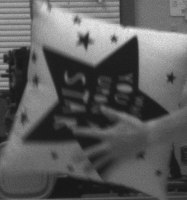}
    }
    \subfigure[0.1 ms]{
  \includegraphics[width=.2\linewidth]{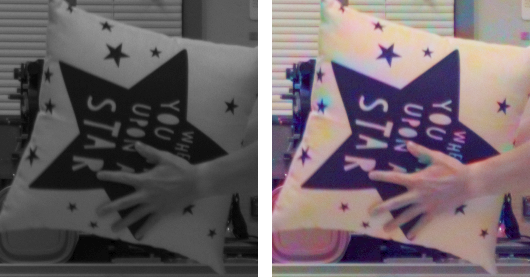}
    }
    \subfigure[4.4 ms]{
  \includegraphics[width=.2\linewidth]{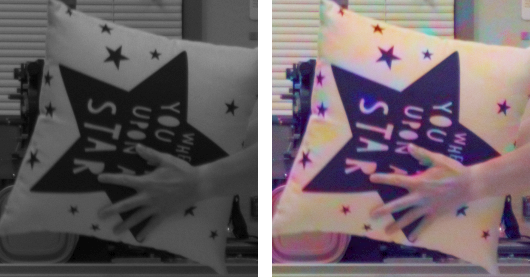}
    }
    \subfigure[8.7 ms]{
  \includegraphics[width=.2\linewidth]{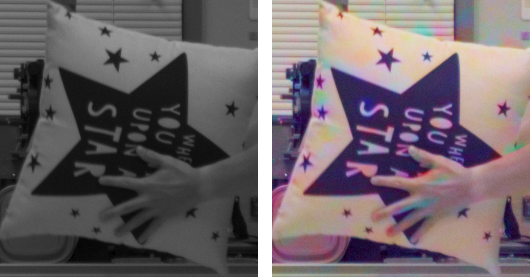}
    }
    \subfigure[13.0 ms]{
  \includegraphics[width=.2\linewidth]{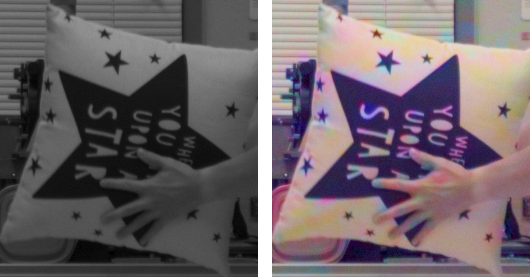}
    }
    \subfigure[$26$ ms]{
  \includegraphics[width=.098\linewidth]{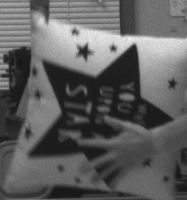}
    }
    \subfigure[17.3 ms]{
  \includegraphics[width=.2\linewidth]{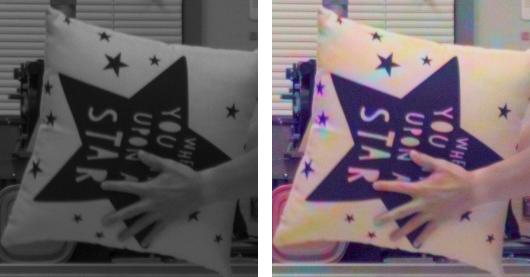}
    }
    \subfigure[21.6 ms]{
  \includegraphics[width=.2\linewidth]{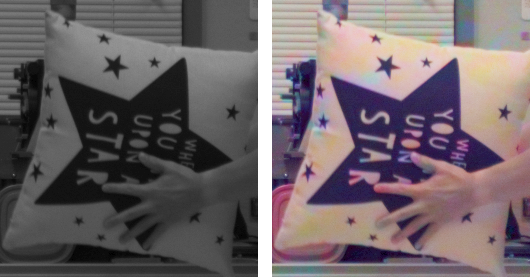}
    }
    \subfigure[25.9 ms]{
  \includegraphics[width=.2\linewidth]{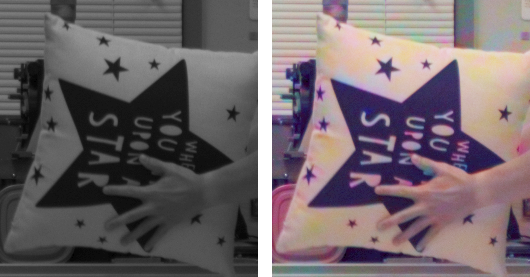}
    }
    \subfigure{
  \includegraphics[width=.2\linewidth]{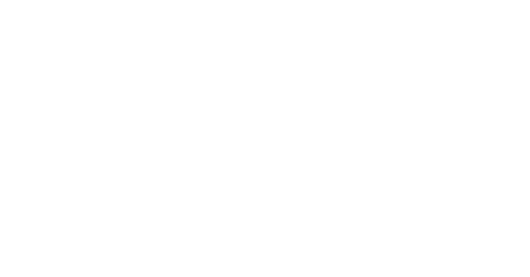}
    }
    \caption{The patches of the moving pillow. (a) and (f) are two consecutive frames from a 600 nm LFR camera. The left images in (b) - (e) and (g) - (i) are the patches from the HFR camera while the right images are from the fused results. The labels are the estimated time elapsed from the moment that (a) was captured. Three LFR frames captured at 0 ms were used to generate the results in (b) - (e), while three LFR frames captured at 26 ms were used to generate the results in (g) - (h). The color balance is adjusted for visual purposes.}
\label{fig:high-low-framerate-result-sequence}
\end{figure*}

Given two sets of LFR frames with three filters (6 frames in total) captured at 0 and 26 ms, PAT fused seven HFR frames with color in between. Figure \ref{fig:high-low-framerate-result-sequence} shows the patches of the moving pillow in the fused results. The pillow in the fused patches was in color with sharp motion boundaries, compared to LFR patches.

\section{Conclusion}
In this paper, we proposed a physics-aware transformer for image fusion on array cameras. This network architecture incorporates tailored receptive fields to reflect the physics of the imaging system. The proposed pipeline of data synthesis effectively provides training data for transformers and has the potential to benefit other learning algorithms. We demonstrated the versatility of PAT on four different camera arrays. We envision PAT being a standard processing tool for array cameras of the next generation, and inspiring designs, combinations and applications of array cameras for better light field sampling.

\bibliographystyle{unsrt}
\bibliography{references}






\end{document}